\documentclass[12pt]{iopart}

\usepackage[bookmarks, colorlinks=true, plainpages = false, citecolor = green, urlcolor = blue, filecolor = blue]{hyperref}
\usepackage{graphicx,subfigure,bm,latexsym}
\usepackage{multirow}
\usepackage{amssymb,bm}
\newcommand{\map}{\mbox{\scriptsize MAP}}

\newcommand{\argmin}{\mbox{argmin}}
\newcommand{\argmax}{\mbox{argmax}}

\begin{document}

\title{Bayesian Optical Flow with Uncertainty Quantification}

\author{Jie Sun}
\address{Department of Mathematics, Clarkson University, Potsdam, New York 13699, USA}
\address{Department of Physics, Clarkson University, Potsdam, New York 13699, USA}
\address{Department of Computer Science, Clarkson University, Potsdam, New York 13699, USA}
\address{Clarkson Center for Complex Systems Science ($C^3S^2$), Clarkson University, Potsdam, New York 13699, USA}
\ead{sunj@clarkson.edu}

\author{Fernando J. Quevedo}
\address{Department of Mechanical and Aerospace Engineering, Clarkson University, Potsdam, New York 13699, USA}
\address{Clarkson Center for Complex Systems Science ($C^3S^2$), Clarkson University, Potsdam, New York 13699, USA}

\author{Erik Bollt}
\address{Department of Mathematics, Clarkson University, Potsdam, New York 13699, USA}
\address{Department of Electrical and Computer Engineering, Clarkson University, Potsdam, New York 13699, USA}
\address{Clarkson Center for Complex Systems Science ($C^3S^2$), Clarkson University, Potsdam, New York 13699, USA}


\begin{abstract}
Optical flow refers to the visual motion observed between two consecutive images. Since the degree of freedom is typically much larger than the constraints imposed by the image observations, the straightforward formulation of optical flow as an inverse problem is ill-posed. Standard approaches to determine optical flow rely on formulating and solving an optimization problem that contains both a data fidelity term and a regularization term, the latter effectively resolves the otherwise ill-posedness of the inverse problem. In this work, we depart from the deterministic formalism, and instead treat optical flow as a statistical inverse problem. We discuss how a classical optical flow solution can be interpreted as a point estimate in this more general framework. The statistical approach, whose ``solution" is a distribution of flow fields, which we refer to as Bayesian optical flow, allows not only ``point" estimates (e.g., the computation of average flow field), but also statistical estimates (e.g., quantification of uncertainty) that are beyond any standard method for optical flow. As application, we benchmark Bayesian optical flow together with uncertainty quantification using several types of prescribed ground-truth flow fields and images.
\end{abstract}


\section{Introduction}
Optical flow reflects the visual motion between consecutive images. 
Determination of optical flow is important for applications ranging from machine learning and computer vision~\cite{SzeliskiBook}, artificial intelligence and robotics~\cite{barron,souhila2007optical}, to scientific applications from oceanography to weather forecasting~\cite{bays,Basnayake2014,cohen,lutt,Luttman2013}, to name a few. In classical approaches optical flow is determined by solving a variational optimization problem which requires inclusion of a regularization term in additional to data fitting for the problem to be well-posed~\cite{Aubert1999,HS1981,Weickert2001}. The choice of regularization parameter turns out critical for the satisfactory inference of optical flow. Despite the many (competing) methods of selecting the regularization parameter, none seems to be most ``natural" comparing to the others~\cite{HansenBook,Thompson,VogelBook}. Another important feature of classical optical flow approaches is that they produce a single solution (by design) as a ``point estimate". In practice, the magnitude of the flow as well as measurement error and noise can vary significantly from one part of the images to another, and thus the at a given pixel the inferred flow vector can be associated with either large or small uncertainty. Although not captured in classical optical flow, such ``uncertainty" would provide valuable information if attainable as part of the solution.

In this work we adopt a statistical inversion framework for the computation of optical flow. In this framework the estimation of optical flow is reformulated from the Bayesian perspective as a statistical inversion problem. From this new, statistical perspective, different types of information black and model assumptions are collectively fused to give rise to a posterior distribution of the unknowns of interest. The posterior distribution, which is typically sampled using some appropriately designed Markov chain Monte Carlo scheme, can be further used to derive various estimates. Importantly, unlike the point estimate of optical flow obtained by classical variational approaches, the proposed statistical inversion approach is a methodological way of uncertainty propagation to produce a distribution of candidate optical flow fields from which various statistical properties can be extracted, including ensemble average estimation and uncertainty quantification. 

The rest of the paper is organized as follows. In Section 2 we review the standard optical flow setup and discuss how it can be treated as an inverse problem defined in finite dimensions. In Section 3 we we review the basics of inverse problems, including the standard least squares approach and Tikhonov regularization. Then, in Section 4 we present a statistical inversion approach for optical flow, and discuss the choice of priors, models, sampling procedure and computational algorithms. In Section 5 we showcase the utility statistical inversion approach for optical flow using several benchmark examples of flow fields and noisy image data. Finally, Section 6 contains a conclusion and discussion of several unsolved and unattempted issues that might be of interest for future research.

\section{Optical Flow as an Inverse Problem}
Given a sequence of consecutively captured images, the visual relative motion between them is commonly referred to as optical flow, which can often provide insight about the actual physical motion. The inference of optical flow is an outstanding scientific question, which requires making assumptions about the underlying motion as well as the measurement process. 

\subsection{Problem Setup} Consider two single-channeled (typically grayscale) digital images (``pictures") taken from the same scene at two nearby time instances. The image data are represented by two matrices $F=[F_{ij}]_{n_x\times n_y}$ and $G=[G_{ij}]_{n_x\times n_y}$. Thus each image contains $n_x\times n_y$ pixels, defined on a common two-dimensional subspace $\Omega$. The goal of optical flow estimation is to infer a flow field, defined by matrices $U=[U_{ij}]_{n_x\times n_y}$ and $V=[V_{ij}]_{n_x\times n_y}$ where $\langle U_{ij},V_{ij}\rangle$ represents the optical ``velocity" occurring at the $(i,j)$-th pixel inferred from the two images. 
The image data $F$ and $G$ are often regarded as sampled data from smooth functions $F(x,y)$ and $G(x,y)$, with $F_{ij}=F(x_i,y_j)$ and $G_{ij}=G(x_i,y_j)$ where $\{(x_i,y_j)\}_{(i=1,\dots,n_x;j=1,\dots,n_y)}$ are grid points from a spatial domain $\Omega$. Thus $U$ and $V$ can be viewed as discrete spatial samples of a smooth 2D flow field $\overrightarrow{W}(x,y)=\langle U(x,y),V(x,y)\rangle$, which is defined on $\Omega$  that captures the visual optical motion occurring between the two observed images.

\subsection{Variational Approach of Inferring Optical Flow}
The classical variational approach of optical flow starts by defining an ``energy" functional whose minimization yields an estimation of the optical flow field~\cite{Aubert1999}.
One of the most widely used functional was proposed by Horn and Schunck in 1981~\cite{HS1981}, given by
\begin{equation}\label{eq:Euv}
E(U,V)=\int \!\!\! \int_{\Omega}(F_x U + F_y V + F_t)^2 dxdy + \alpha\int \!\!\! \int_{\Omega}(\|\nabla U\|^2+\|\nabla V\|^2)dxdy,
\end{equation}
where $U(x,y)$ and $V(x,y)$ are smooth functions defined over $\Omega$ which represent a candidate flow field. In the Horn-Schunck functional, the first term is often referred to as {\it data fidelity} as it measures the deviation of the total image intensity from being conservative, that is, how much does the model fit deviates from the condition 
\begin{equation}\label{eq:cons}
	dF/dt=F_x U + F_y V + F_t = 0.
\end{equation}
The second term measures the {\it solution regularity} by penalizing solutions that have large spatial gradients and is called the regularization term. The relative emphasis of smoothness as compared to ``fitting" the total image intensity conservation equation~(\ref{eq:cons}) is controlled by the positive scalar $\alpha$ which is called a {\it regularization parameter}. The main role of the regularization term is to ensure that the minimization of the functional is a well posed problem. Without the regularization term, the problem is ill-posed.

Given $\alpha$, the functions $U$ and $V$ that minimize the Horn-Schunck functional~(\ref{eq:Euv}) satisfy the Euler-Lagrange equations
\begin{equation}
\left\{
\begin{array}{lr}
        F_x(F_xU+F_yV+F_t) = \alpha(U_{xx}+U_{yy}),\\
        F_y(F_xU+F_yV+F_t) = \alpha(V_{xx}+V_{yy}),
        \end{array}
\right.
\end{equation}
which are typically solved by some iterative scheme over a finite set of spatial grid points~\cite{HS1981,Tarnec2014} to produce an estimation of the optical flow. Alternatively, one could also discretize the functional~(\ref{eq:Euv}) itself to yield a finite-dimensional inverse problem as discussed in Section 2.3 below with solution strategy reviewed in Section~\ref{Sec3}.

\subsection{Finite-dimensional representation of the Variational Optical Flow Functional}
As discussed in the previous section, the classical variational approach of optical flow works by first formulating and minimizing a functional over smooth vector fields, and then evaluating the obtained vector field at the grid points on which the original image data are given. Here we approach the problem from a different route, by first discretizing the functional~(\ref{eq:Euv}) to convert the functional minimization (an infinite-dimensional problem) into a finite-dimensional linear inverse problem, and then solving the inverse problem to yield solutions which give the values of a vector field defined over the same grid points as the image data.

The remainder of this section will be focused on the conversion of the functional~(\ref{eq:Euv}) into a finite-dimensional function defined on a set of uniformly distributed grid points
\begin{equation}
\{(x_i,y_j)\}_{i=1,2,\dots,n_x; j=1,2,\dots,n_y},
\end{equation}
where $x_{i+1}-x_i=\Delta{x}$ and $y_{j+1}-y_j=\Delta{y}$ are the spacing in the $x$ direction and the $y$ direction, respectively. The conversion will be achieved by approximating the integrals in Eq.~(\ref{eq:Euv}) with appropriately derived summations over the grid points. For notational convenience, we use a bold lowercase variable to denote the vectorization of a matrix. For example, the boldface vector $\bm{q}$ denotes the column vector obtained by ``vertically stacking" the columns of a matrix $Q=[\overrightarrow{Q}_1,\overrightarrow{Q}_2,\dots,\overrightarrow{Q}_n]$ in order~\cite{GolubBook}, where $\overrightarrow{Q}_i$ denotes the $i$-th column of $Q$. That is,
\begin{equation}
	\bm{q}=\mbox{vec}(Q)=
  \left[ {\begin{array}{c}
   \overrightarrow{Q}_1\\
   \overrightarrow{Q}_2\\
   \vdots\\
   \overrightarrow{Q}_n   
  \end{array} } \right].
\end{equation}

First let us consider the data fidelity term: $\int \!\!\! \int_\Omega(F_xU+F_yV+F_t)^2dxdy$. The spatial derivatives $F_x(x,y)$ and $F_y(x,y)$ can be approximated by a finite difference scheme. For example, the simple {\it forward difference} yields the approximations:
\begin{equation}
\left\{
\begin{array}{lr}
F_x(x,y)\approx\frac{1}{\Delta{x}}\left[F(x+\Delta{x},y)-F(x,y)\right],\\
F_y(x,y)\approx\frac{1}{\Delta{y}}\left[F(x,y+\Delta{y})-F(x,y)\right].
        \end{array}
\right.
\end{equation}
We next express these derivatives as operations on the column vector $\bm{f}$. To do this, we define matrix $S_k=[S_k^{(ij)}]_{k\times k}$ as
\begin{equation}
S_k^{(ij)}=
\left\{
\begin{array}{lr}
-\delta_{ij} + \delta_{i+1,j}, & \mbox{~if $i<k$;}\\
-\delta_{i,j-1} + \delta_{i,j}, & \mbox{~if $i=k$}.
        \end{array}
\right.
\end{equation}
The forward difference applied to $\bm{f}$ can be represented as
\begin{equation}
\left\{
\begin{array}{lr}
\bm{f}_x\approx Q_x\bm{f},\\
\bm{f}_y\approx Q_y\bm{f},
        \end{array}
\right.
\end{equation}
where
\begin{equation}\label{eq:Qxy}
\left\{
\begin{array}{lr}
Q_x\equiv\frac{1}{\Delta{x}}[I_n\otimes S_m],\\
Q_y\equiv\frac{1}{\Delta{y}}[S_n\otimes I_m].
\end{array}\right.
\end{equation}
The temporal derivative can be estimated from the data by the direct pixel-wise difference between the two images, to yield
\begin{equation}\label{eq:ft}
\bm{f}_t\approx \bm{g}-\bm{f}.
\end{equation}

With these definitions and approximations, we obtain a discretized version of the conservation equation~(\ref{eq:cons}) expressed as a finite-dimensional linear system:
\begin{equation}
A\bm{x}=\bm{b},
\end{equation}
where
\begin{equation}
\left\{
\begin{array}{lr}
A = [\mbox{diag}(\bm{f}_x), \mbox{diag}(\bm{f}_y)],\\
\bm{x} = [\bm{u}^\top,\bm{v}^\top]^\top,\\
\bm{b} = -\bm{f}_t.
\end{array}\right.
\end{equation}
Here $\mbox{diag}(\bm{f}_x)$ and $\mbox{diag}(\bm{f}_y)$ represent the diagonal matrices whose diagonal elements are given by the entries of the vector $\bm{f}_x$ and $\bm{f}_y$, respectively.
From this connection we approximate the first integral in the functional~(\ref{eq:Euv}) as $\|A\bm{x}-\bm{b}\|^2$ where $\|\cdot\|$ denotes the standard Euclidean norm.

Next we develop a finite-dimensional approximation of the regularization term in the functional~(\ref{eq:Euv}). This requries discretization of $\nabla{U}$ and $\nabla{V}$. Using a similar forward difference to approximate the partial derivatives, we obtain
\begin{equation}
\left\{
\begin{array}{lr}
\nabla U\approx \frac{1}{\Delta{x}}\left[U(x+\Delta{x},y)-U(x,y)\right] + \frac{1}{\Delta{y}}\left[U(x,y+\Delta{y})-U(x,y)\right],\\
\nabla V\approx \frac{1}{\Delta{x}}\left[V(x+\Delta{x},y)-V(x,y)\right] + \frac{1}{\Delta{y}}\left[V(x,y+\Delta{y})-V(x,y)\right].
\end{array}\right.
\end{equation}
For the vectorized variables $\bm{u}$ and $\bm{v}$, we have
\begin{equation}
\left\{
\begin{array}{lr}
\nabla\bm{u}=\bm{u}_x+\bm{u}_y\approx(Q_x+Q_y)\bm{u},\\
\nabla\bm{v}=\bm{v}_x+\bm{v}_y\approx(Q_x+Q_y)\bm{v},
\end{array}\right.
\end{equation}
where $Q_x$ and $Q_y$ are defined in Eq.~(\ref{eq:Qxy}). Consequently, we obtain the approximation of the second integral in the Horn-Schunck functional~(\ref{eq:Euv}) as
\begin{equation}
\left\{
\begin{array}{lr}
\int \!\!\! \int_{\Omega}\|\nabla{U}\|^2dxdy\approx\bm{u}_x^\top\bm{u}_x+\bm{u}_y^\top\bm{u}_y\approx\bm{u}^\top[Q_x^\top Q_x+Q_y^\top Q_y]\bm{u},\\
\int \!\!\! \int_{\Omega}\|\nabla{V}\|^2dxdy\approx\bm{v}_x^\top\bm{u}_y+\bm{v}_y^\top\bm{v}_y\approx\bm{v}^\top[Q_x^\top Q_x+Q_y^\top Q_y]\bm{v},
\end{array}\right.
\end{equation}
which then gives
\begin{equation}
	\int \!\!\! \int_{\Omega}(\|\nabla u\|^2+\|\nabla v\|^2)dxdy\approx \bm{x}^\top Q\bm{x},
\end{equation}
where the matrix
\begin{equation}\label{eq:Q}
	Q=I_2\otimes[Q_x^\top Q_x+Q_y^\top Q_y].
\end{equation}

Therefore, the variational optical flow formulation~(\ref{eq:Euv}) can be reformulated at a finite spatial resolution as an inverse problem, where, for a given parameter value $\alpha$, the corresponding solution is given by solving the following (regularized) optimization:
\begin{equation}
	\min_{\bm{x}}\left(\|A\bm{x}-\bm{b}\|^2 + \alpha\bm{x}^\top Q\bm{x}\right).
\end{equation}
This is a standard approach in inverse problems, formulated as a least squares problem with Tikhonov regularization, with more details to be presented in the next section.

\section{Inverse Problem in Finite Dimensions}\label{Sec3}
Although there has been a great deal of  progress on the mathematical characterization of inverse problems in the field of functional analysis, a practical problem often concerns finding a solution in a finite-dimensional space. At a fundamental level, the most common inverse problem stems from a linear model~\cite{HansenBook,KaipioBook,VogelBook}
\begin{equation}\label{eq:Axb}
\bm{b}=A\bm{x}+\bm{\eta},
\end{equation}
where $\bm{b}\in\mathbb{R}^m$ is a column vector of {\it observed data}, $A=[a_{ij}]_{m\times n}\in\mathbb{R}^{m\times n}$ is a (known)  matrix representing the underlying model, the column vector $\bm{\eta}\in\mathbb{R}^n$ denotes (additive) noise, and $\bm{x}\in\mathbb{R}^n$ is the vector of {\it unknowns} to be inferred. 

Given $A$ and $\bm{b}$, the problem of inferring $\bm{x}$ in Eq.~(\ref{eq:Axb}) is called an {\it inverse problem} because rather than direct ``forward" computation from the model, it requires a set of indirect, ``backward", or ``inverse" operations to determine the unknowns~\cite{VogelBook}.   Depending on the rank and conditioning of the matrix $A$, the problem may be ill-posed or ill-conditioned. In classical approaches, these issues are dealt with by adjusting the original problem to a (slightly) modified optimization problem as discussed in  Section~\ref{ls} whose solution is meant to represent the original, as discussed in Section~\ref{tik}.

We note that in the classical setting a solution to the inverse problem is a vector $\bm{x}$ as a result of solving an optimization problem. Such a solution is referred to as an {\it point estimate} because it gives one solution vector without providing any information about how reliable (or uncertain) the solution is~\cite{HansenBook,VogelBook}. On the other hand, the statistical inversion approach to inverse problems provides an {\it ensemble} of solutions---defined by the posteriori distribution which not only point estimates can be made but also their uncertainty quantification~\cite{GelmanBook,KaipioBook}.

\subsection{Least squares solution}\label{ls}
The classical least squares solution to the inverse problem is given by~\cite{GolubBook}
\begin{equation}
	\bm{x}_{\ell_2} = A^\dagger{\bm{b}},
\end{equation}
where $A^\dagger$ denotes the pseudo-inverse of $A$ which can be obtained from the singular value decomposition of $A$~\cite{GolubBook}.
Depending on the rank of $A$, the least squares solution $\bm{x}_{\ell_2}$ is associated with one of the minimization problems:
\begin{equation}
\left\{
\begin{array}{lr}
\min_{A\bm{x}=\bm{b}}\|\bm{x}\|_2, & \mbox{if $\mbox{rank}(A)< n$;}\\
\min_{\bm{x}}\|A\bm{x}-\bm{b}\|_2, & \mbox{if $\mbox{rank}(A)= n$.}
\end{array}\right.
\end{equation}
Here $\|\cdot\|_2$ denotes the $\ell_2$ (Euclidean) norm.
Let the true solution to Eq.~(\ref{eq:Axb}) be $\bm{x}^*$, that is, $\bm{b}=A\bm{x}^*+\bm{\eta}$. It follows that
\begin{equation}
	\bm{x}_{\ell_2}-\bm{x}^* = A^\dagger\bm{\eta}.
\end{equation}
In practice, even when the matrix $A$ has full column rank ($\mbox{rank}(A)=n$), the discrepancy between the true and least squares solutions is typically dominated by noise when some singular values of $A$ are close to zero, rendering $A$ an ill-conditioned matrix and the solution $\bm{x}^*$ unstable and sensitive to small changes in data~\cite{VogelBook}.

\subsection{Tikhonov Regularization}\label{tik}
A powerful approach to resolve the instabilities due to noise and the near-singularity of $A$ is to {\it regularize} the problem. In the classical Tikhonov regularization one adds a quadratic regularization term $\alpha\bm{x}^TL\bm{x}$ for some prescribed matrix $L$ to penalize non-smoothness, giving rise to a regularized optimization problem~\cite{Tikhonov1963,Tikhonov1977,Tikhonov1990,VogelBook}:
\begin{equation}\label{eq:treg0}
	\min_{\bm{x}}\left(\|A\bm{x}-\bm{b}\|_2^2 + \alpha\bm{x}^TL\bm{x}\right).
\end{equation}
In the regularized problem the positive scalar parameter $\alpha$ controls the weight of regularization and $L$ is typically a symmetric positive definite matrix, both of which need to be chosen appropriately for the problem to be uniquely defined~\cite{Tikhonov1963,Tikhonov1977,Tikhonov1990,VogelBook}.  We refer to the two terms $\|A\bm{x}-\bm{b}\|_2^2$ and $\alpha\bm{x}^\top L\bm{x}$ in~(\ref{eq:treg0}) as {\it data fidelity} and {\it solution regularity}, respectively. In a simplistic description, they can be described as ``selecting" a solution $\bm{x}_{\alpha}$ that balances the desire to ``solve" $A\bm{x}=\bm{b}$ and to be ``regular" as measured by $\bm{x}^\top L\bm{x}$. The regularization parameter $\alpha$ therefore dictates the extent to which the compromise is made between the two.

For a given parameter $\alpha$, we denote the corresponding regularized solution by
\begin{equation}\label{eq:treg}
	\bm{x}_{\alpha}=\argmin_{\bm{x}}\{\|A\bm{x}-\bm{b}\|^2 + \alpha\bm{x}^\top L\bm{x}\}.
\end{equation}
By standard vector calculus, it can be shown that $\bm{x}_{\alpha}$ is in fact a solution to the modified linear system 
\begin{equation}\label{eq:treg2}
	(A^\top A+\alpha L)\bm{x}_{\alpha} = A^\top\bm{b},
\end{equation}
which is typically well-posed for appropriate choices of $L$ and $\alpha$.
When the matrices are large and sparse, Eq.~(\ref{eq:treg2}) is often solved by iterative methods rather than a direct matrix inversion since the latter tends to be numerically costly and unstable~\cite{GolubBook}.

In Tikhonov regularization, a key issue is how to choose the regularization parameter $\alpha$ appropriately. In theory, a ``good" regularizer has the property that in the absence of noise, the solution to the regularized problem converges to the true solution when the regularization parameter $\alpha\rightarrow0$. However, in practice, under the presence of noise, it is always a challenge to try to determine a good value for $\alpha$. If $\alpha$ is too small, the instability and sensitivity of the original problem would still persist; whereas for too large of an $\alpha$ the solution will be over-regularized and not fit the data well. A good balance is thus paramount. Despite the existence and ongoing development of many competing methods for selecting $\alpha$ most of which focus on asymptotical optimality as the number of data points approach infinity, none of them stands out as \textcolor{black}{a best} ``natural" choice unless specific priori information about the noise in the data are available (see Chapter 7 of Ref.~\cite{VogelBook}). In the following section we discuss how the problem of selecting an exact regularization parameter is no longer required from a Bayesian perspective; instead, it suffices to start with some loose range of values represented by a probability distribution, unless more specific knowledge is available about the problem in which case the distribution can be chosen to reflect such information.

\section{Statistical Inversion Approach}
The statistical inversion approach to an inverse problem starts by treating all variables as random variables (e.g., $\bm{x}$, $\bm{b}$ and $\bm{\eta}$ in Eq.~(\ref{eq:Axb})), and representing our knowledge (or absence of knowledge) of the unknowns as prior distributions. With observational data and a forward model (e.g., matrix $A$ in Eq.~(\ref{eq:Axb})), the ``inversion" leads to an ensemble of points together with a posterior distribution from which various statistical estimates can be made~\cite{OSullivan1986,GelmanBook,KaipioBook,Cavalier2008}.
The key in the inversion is to use the Bayes rule to express the {\it posterior} distribution $p(\bm{x}|\bm{b})$, which is the conditional distribution of the ``solution vector" $\bm{x}$ given the observed data $\bm{b}$,  as~\cite{GelmanBook,KaipioBook}
\begin{equation}
p(\bm{x}|\bm{b})=\frac{1}{p(\bm{b})}p(\bm{b}|\bm{x})\cdot p(\bm{x}).
\end{equation}
Here the {\it likelihood} function $p(\bm{b}|\bm{x})$ is the probability density function (pdf) of the random variable $\bm{b}$ given $\bm{x}$ which is determined by the underlying model;
$p(\bm{x})$ is the {\it priori} distribution of $\bm{x}$;
and $p(\bm{b})>0$ acts as a normalization constant which does not affect the solution procedure or the final solution itself.

Thus, in the statistical inversion formulation, each candidate solution $\bm{x}$ is associated with the probability $p(\bm{x}|\bm{b})$ that is determined (up to a normalization constant $1/p(\bm{b})$) once the likelihood function and the prior distribution are given. For a given inverse problem, the likelihood function can be obtained using the underlying model such as Eq.~(\ref{eq:Axb}) including the noise distribution. On the other hand, the prior distribution $p(\bm{x})$ is typically constructed according to some prior knowledge of the solution.  Unlike standard approaches which produce ``point estimates", in statistical inversion it is the posterior distribution that is considered as the ``solution" to the inverse problem. Based on the posterior distribution, one can further extract useful information such as point estimates and uncertainty quantification, by sampling from the distribution. Efficient sampling methods will be reviewed toward the end of this section.

The unique feature of enabling information fusion and uncertainty quantification has made the statistical inversion approach to inverse problems an attractive venue for the development of new theory and applications. In image processing applications, it has been utilized for many problems such as image denoising and deblurring~\cite{Bardsley2012,Lebrun2013}, sparse signal reconstruction~\cite{Seeger2011}, and more recently attempted for optical flow computation~\cite{Chantas2014}. In particular, we note that our approach, although different in many of the technical aspects, shares a similar statistical inversion perspective as Ref.~\cite{Chantas2014}.

\subsection{Statistical interpretation of optical flow obtained by Tikhonov regularization}
Under the statistical inversion framework, solution to an inverse problem is the posterior distribution, from which point estimates can be further obtained. Among these point estimates, a particularly common one is the {\it maximum a posteriori} (MAP) estimator, which is defined as
\begin{eqnarray}
\bm{x}_{\map} &=& \argmax_{\bm{x}}p(\bm{x}|\bm{p}) = \argmin_{\bm{x}}\{-\ln p(\bm{x}|\bm{b})\} \\
		&=& \argmin_{\bm{x}}\{-\ln p(\bm{b}|\bm{x}) - \ln p(\bm{x})\}.\label{eq:xmap}
\end{eqnarray}
As noted in Refs.~\cite{Bardsley2012,KaipioBook}, the MAP estimator given by Eq.~(\ref{eq:xmap}) produces a vector $\bm{x}$ that is identical to the solution of a Tikhonov regularization specified in Eq.~(\ref{eq:treg}) upon appropriate choice of the model and prior pdf. In particular, consider the model given by Eq.~(\ref{eq:Axb}) with independent and identically distributed (iid) Gaussian noise of variance $\lambda^{-1}$. It follows that
\begin{equation}
	p(\bm{b}|\bm{x})=p(\bm{\eta})\propto \exp\left(-\frac{\lambda}{2}\|A\bm{x}-\bm{b}\|^2\right),
\end{equation}
where the symbol ``$\propto$" means ``proportional to". Under a Gaussian prior,
\begin{equation}
	p(\bm{x})\propto\exp\left(-\frac{\delta}{2}\bm{x}^\top L\bm{x}\right),
\end{equation}
the term $-\ln p(\bm{x}|\bm{b})$ in the MAP estimator becomes
\begin{equation}\label{eq:lnmap}
	-\ln p(\bm{x}|\bm{b}) \propto \|A\bm{x}-\bm{b}\|^2+(\delta/\lambda)\bm{x}^\top L\bm{x}.
\end{equation}
The choice of $\alpha=\delta/\lambda$ in the Tikhonov regularization then yields a solution $\bm{x}_{\alpha}$ that equals the vector $\bm{x}_{\map}$ given by the same MAP estimator. Thus, with these assumptions of the form of the noise, the distribution of the prior, there is a logical bridge between two different philosophies for inverse problems, in that a solution from Tikhonov regularization can be interpreted as an MAP estimate of the posteriori distribution under appropriate choices of prior distributions and likelihood functions.

Furthermore, and crucially, as shown in the next section, more important information exists in the statistical inversion framework.  Specifically, by sampling from the posterior distribution the statistical inversion approach allows for not only a point estimate but also other statistical properties associated with the solution, in particular spread around a point estimate which enables uncertainty quantification.

\subsection{Choice of Priors and Hyperpriors}
In general, the statistical inversion framework allows flexibility in choosing prior distributions about unknowns and noise. When these prior distributions themselves contain unknown parameters, these parameters can themselves be thought of as random variables which follow hyperprior distributions. In this paper we will consider some elementary choice of the priors and hyperpriors, with the main purpose of showing how to setup the procedure for sampling the posterior distribution.

We focus on (multivariate) Gaussian priors for both the unknown $\bm{x}$ and noise $\bm{\eta}$. In the absence of knowledge of how ``spread-out" the prior distributions are, we use additional parameters for these priors, where these parameters are taken to be random variables drawn from hyperprior distributions. This type of approach has been previously adopted and implemented in image denoising applications~\cite{Bardsley2012}. In particular, we consider a specific class of the noise and prior distributions:
\begin{equation}
\left\{
\begin{array}{lr}
\mbox{likelihood function:~}p(\bm{b}|\bm{x},\lambda)\propto\lambda^{m/2}\exp\left(-\frac{\lambda}{2}\|A\bm{x}-\bm{b}\|^2\right),\\
\mbox{prior distribution:~}p(\bm{x}|\delta)\propto\delta^{n/2}\exp\left(-\frac{\delta}{2}\bm{x}^\top L\bm{x}\right).\\
\end{array}\right.
\end{equation}
Here the noise is assumed to be additive, Gaussian, and independent of the measured data, with variance $\lambda^{-1}$, giving rise to the form of the likelihood function. On the other hand, the prior distribution is considered to be Gaussian with covariance matrix $(\delta L)^{-1}$ (matrix $\delta L$ is referred to as the {\it precision} matrix). 
We choose $L=Q$ where $Q$ is given by Eq.~(\ref{eq:Q}) which corresponds to a spatial regularization measure. As it turns out, this choice of $L$ is closely related to the selection of prior according to a spatial Gaussian Markov random field which is common in tackling spatial inverse problems~\cite{Bardsley2013,Higdon}.

As we pointed out, to completely specify the posterior distribution, we also need to choose prior distributions for the parameters $\lambda$ and $\delta$. These are often called {\it hyperpriors}. Following Ref.~\cite{Bardsley2012}, we choose the priors for $p(\lambda)$ and $p(\delta)$ to be Gamma distributions, as
\begin{equation}
\left\{
\begin{array}{lr}
p(\lambda)\propto\lambda^{\alpha_\lambda-1}\exp(-\beta_\lambda\lambda),\\
p(\delta)\propto\delta^{\alpha_\delta-1}\exp(-\beta_\delta\delta).\\
\end{array}\right.
\end{equation}
Such choice ensures that $p(\lambda)$ and $p(\delta)$ are conjugate hyper-priors, and makes the design of sampling of posterior distributions more convenient.
In the absence of knowledge of the values of $\lambda$ and $\delta$, one needs to choose the values of $\alpha_\lambda$, $\beta_\lambda$, $\alpha_\delta$, and $\beta_\delta$ to ensure the distributions $p(\lambda)$ and $p(\delta)$ to be ``wide", allowing the Markov chain to explore a potentially larger region of the parameter space. Following Ref.~\cite{Bardsley2012,Bardsley2013,Higdon}, we set $\alpha_\lambda=\alpha_\delta=1$ and $\beta_\lambda=\beta_\delta=10^{-4}$ unless otherwise noted. We tested other choice of parameters as well and they mainly affect the length of the transient in the MCMC sampling process and do not seem to have a strong influence on the asymptotic outcome. Details of these will be discussed in Section 5 along with numerical examples.

\subsection{Sampling from the Posterior Distribution}
In the statistical inversion formalism, once the form of the posterior distribution is derived, the remaining part of the work is devoted to efficient sampling from the posterior distribution. Typically a Markov chain Monte Carlo (MCMC) sampling approach is adopted. The main idea is to generate a sequence of samples according to a prescribed Markov chain whose unique stationary distribution is the desired posterior distribution.

Note that under the Gaussian priors and gamma hyperpriors as we discussed above, the full conditional distributions that relate to the posterior distribution are given by
\begin{equation}
\left\{
\begin{array}{lr}
p(\bm{x}|\lambda,\delta,\bm{b})\propto\exp\left(-\frac{\lambda}{2}\|A\bm{x}-\bm{b}\|^2-\frac{\delta}{2}\bm{x}^\top L\bm{x}\right),\\
p(\lambda|\bm{x},\delta,\bm{b})\propto\lambda^{m/2+\alpha_\lambda-1}
\exp\left(\lambda\left[-\frac{1}{2}\|A\bm{x}-\bm{b}\|^2+\beta_\lambda\right]\right),\\
p(\delta|\bm{x},\lambda,\bm{b})\propto\delta^{n/2+\alpha_\delta-1}
\exp\left(\delta\left[-\frac{1}{2}\bm{x}^\top L\bm{x} - \beta_\delta\right]\right).
\end{array}\right.
\end{equation}
In other words,
\begin{equation}
\left\{
\begin{array}{lr}
\bm{x}|\lambda,\delta,\bm{b}\sim\mathcal{N}\left((\lambda A^\top A+\delta L)^{-1}\lambda A^\top\bm{b},
(\lambda A^\top A + \delta L)^{-1}\right),\\
\lambda|\bm{x},\delta,\bm{b}\sim\Gamma\left(m/2+\alpha_\lambda,\frac{1}{2}\|A\bm{x}-\bm{b}\|^2+\beta_\lambda\right),\\
\delta|\bm{x},\lambda,\bm{b}\sim\Gamma\left(n/2+\alpha_\delta,\frac{1}{2}\bm{x}^\top L\bm{x}+\beta_\delta\right).
\end{array}\right.
\end{equation}

We adopt the block Gibbs sampler developed in Refs.~\cite{Bardsley2012,KaipioBook} as a specific MCMC procedure to sample the posterior distribution. In theory the sample distribution asymptotically converges to the true posterior distribution. The approach contains the following steps.
\begin{enumerate}
\setcounter{enumi}{0}
\item Initialize $\delta_0$ and $\lambda_0$, and set $k=0$.
\item Sample $\bm{x}^{k}\sim\mathcal{N}\left((\lambda A^\top A+\delta L)^{-1}\lambda A^\top\bm{b},
(\lambda A^\top A + \delta L)^{-1}\right)$.
\item Sample $\lambda_{k+1}\sim\Gamma\left(m/2+\alpha_\lambda,\frac{1}{2}\|A\bm{x}^k-\bm{b}\|^2+\beta_\lambda\right)$.
\item Sample $\delta_{k+1}\sim\Gamma\left(n/2+\alpha_\delta,\frac{1}{2}(\bm{x}^k)^\top L(\bm{x}^k)+\beta_\delta\right)$.
\item Set $k\leftarrow k+1$ and return to Step (ii).
\end{enumerate}
Here the computational burden is mainly due to Step 1, which requires drawing samples from  a multivariate Gaussian variable, which is equivalent to solving the following linear system at each iteration for $\bm{x}^{k}$:
\begin{equation}\label{eq:GibbsIteration}
(\lambda_k A^\top A + \delta_k L)\bm{x}^k = \lambda_k A^\top\bm{b} + \bm{w},
~\mbox{where}~\bm{w}\sim\mathcal{N}(\bm{0},\lambda_k A^\top A+\delta_k L).
\end{equation}
For large matrices, instead of a direct solve using Gauss elimination, an iterative method is usually preferred.
Among the various notable iterative methods such as Jacobi, Gauss-Seidel (G-S), and conjugate gradient (CG)~\cite{SaadBook}, we adopted the CG for all the numerical experiments as reported in this paper, with a starting vector of all zeros, maximum of $500$ iterations, and error tolerance of $10^{-6}$.
Note that the computational bottleneck in solving Eq.~(\ref{eq:GibbsIteration}) can in principle be tackled using more efficient methods, for example, by exploring history of solutions with similar parameters to provide ``good" initial guess or by utilizing the Karhunen-Lo\`{e}ve expansion to reduce the dimensionality of the problem.

\section{Results: Bayesian Optical Flow from Statistical Inversion}
\subsection{Benchmark Flow Fields and Noisy Image Pairs}\label{sec:5.1}
To benchmark the proposed statistical inversion approach for Bayesian optical flow, we consider $5$ qualitatively different flow fields that span a broad diversity of possibilities. The flow fields, which are all defined on the normalized spatial domain $[-1,1]\times[-1,1]$, are generally not divergence-free. In particular, the flow fields are defined as follows.

\smallskip\noindent
{\bf Flow field 1:} $\langle U(x,y),V(x,y)\rangle = \langle x,y\rangle$.

\smallskip\noindent
{\bf Flow field 2:} $\langle U(x,y),V(x,y)\rangle = \langle -y,x\rangle$.

\smallskip\noindent
{\bf Flow field 3:} $\langle U(x,y),V(x,y)\rangle = \langle y,\sin(x)\rangle$.

\smallskip\noindent
{\bf Flow field 4:} $\langle U(x,y),V(x,y)\rangle = \langle -\pi\sin(0.5\pi x)\cos(0.5\pi y),\pi\cos(0.5\pi x)\sin(0.5\pi y)\rangle$.

\smallskip\noindent
{\bf Flow field 5:} $\langle U(x,y),V(x,y)\rangle = \langle -\pi\sin(\pi x)\cos(\pi y),\pi\cos(\pi x)\sin(\pi y)\rangle$.

We first consider synthetic image pairs, where the first image is constructed by the equation
\begin{equation}\label{eq:firstimage}
	F(x,y)=\frac{1}{2}\left[\cos(\pi x)\cos(\pi y)+1\right].
\end{equation}
Then, for each optical flow field, we generate the second image $G$ using the equation
\begin{equation}\label{eq:hsexact}
	\bm{g} = \bm{f} - \bm{f}_x\bm{u} - \bm{f}_y\bm{v} + \bm{\eta},
\end{equation}
where $\bm{f}$ and $\bm{g}$ represent the vectorization of $F$ and $G$, respectively, $\bm{\eta}$ denotes (multivariate) noise whose individual components are independently drawn from a Gaussian distribution with zero mean and fixed standard deviation $\sigma$, and spatial derivatives are numerically implemented using the forward difference scheme. All synthetic images are constructed at the fixed resolution of $30$-by-$30$ pixels, with uniform spacing in both directions, represented by a set of $30$-by-$30$ matrices.

\subsection{Bayesian Optical Flow with Uncertainty Quantification}\label{sec:5.2}
\begin{figure}[htbp]
\centering
\includegraphics[width=0.75\textwidth]{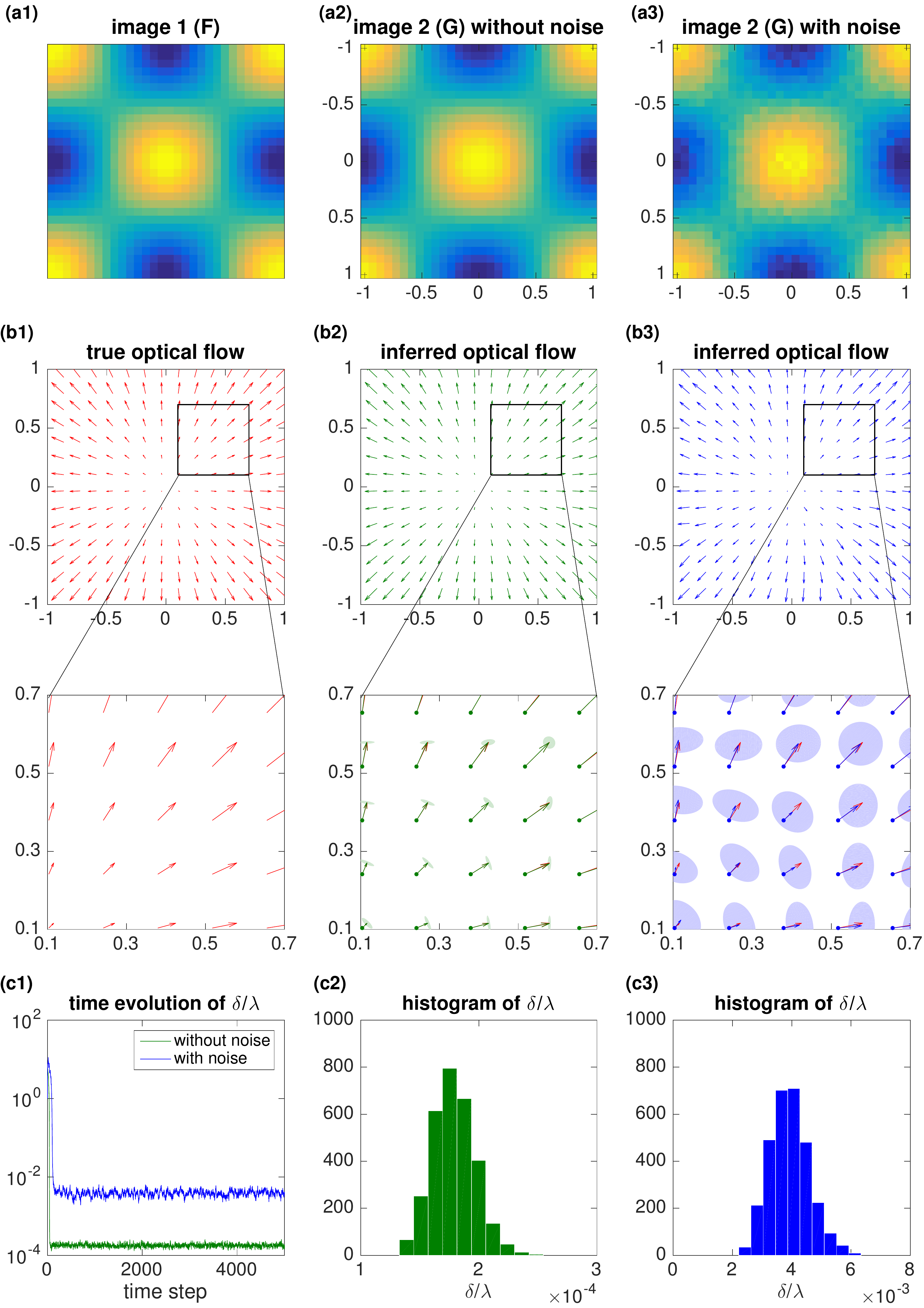}
\caption{Bayesian optical flow computed using statistical inversion, based on image pairs $(F,G)$ under example flow field 1, where $\langle U(x,y),V(x,y)\rangle = \langle x,y\rangle$. Top row (a1-a3): image data generated according to Eq.~(\ref{eq:hsexact}) for $F$ (a1) and using Eq.~(\ref{eq:firstimage}) for $G$ either without noise (a2) or with noise \textcolor{black}{standard deviation $\sigma=0.02$} (a3). Middle rows (b1-b3): true flow field (b1) and the mean flow field from the posterior distribution estimated using the MCMC samples. In the panels below (b2) and (b3), we also show the uncertainty regions (as shaded ellipses), also from MCMC samples as given by Eq.~(\ref{eq:cr}). Bottom row (c1-c3): time evolution (c1) as well as the distribution (c2-c3) of the effective regularization parameter $\sigma/\lambda$, where the distributions are obtained after discarding the initial transient in the MCMC sampling process.}
\label{fig1}
\end{figure}

\begin{figure}[htbp]
\centering
\includegraphics[width=0.75\textwidth]{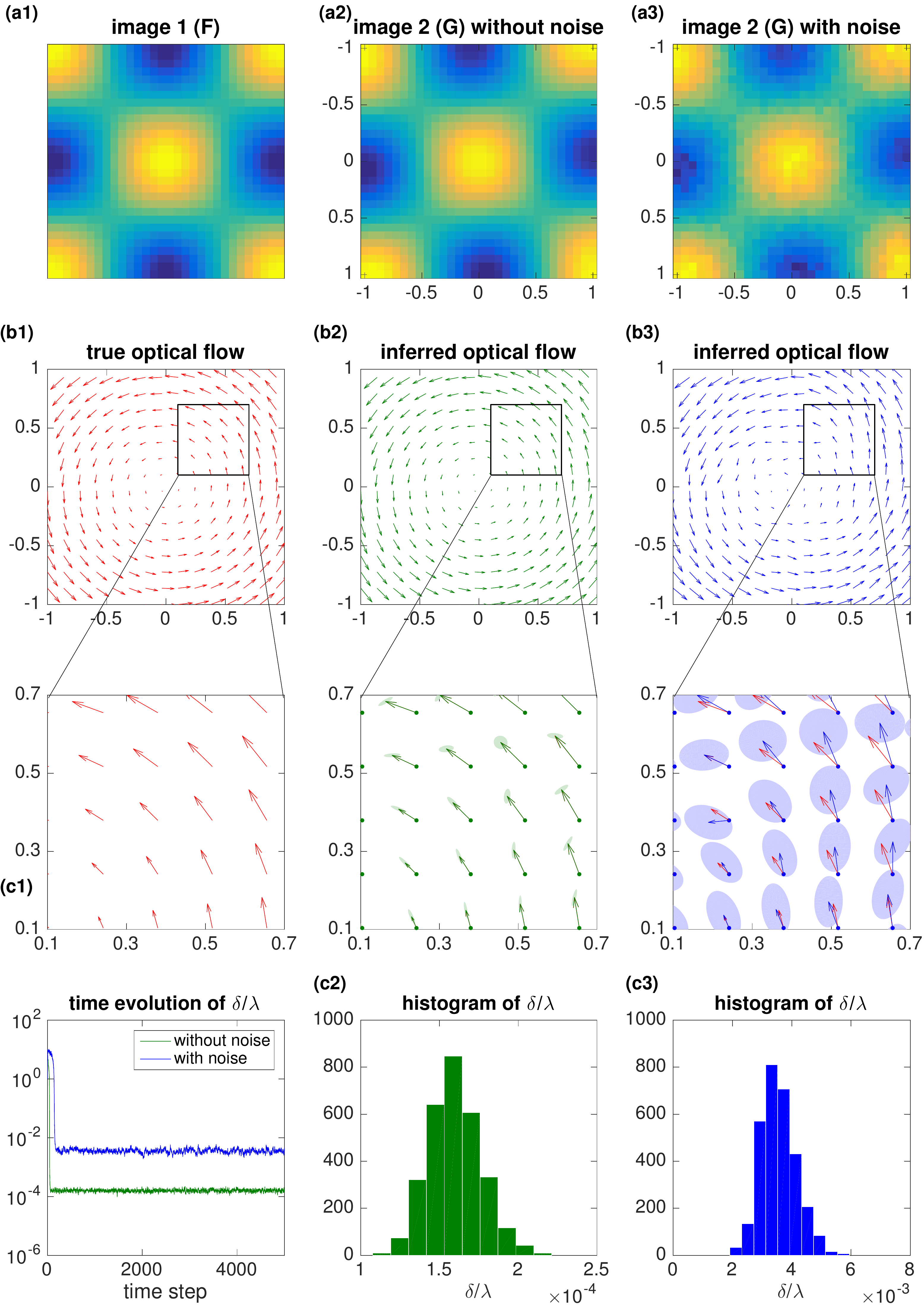}
\caption{Bayesian optical flow computed using statistical inversion, based on image pairs $(F,G)$ under example flow field 2, where $\langle U(x,y),V(x,y)\rangle = \langle -y,x\rangle$.  Top row (a1-a3): image data generated according to Eq.~(\ref{eq:hsexact}) for $F$ (a1) and using Eq.~(\ref{eq:firstimage}) for $G$ either without noise (a2) or with noise \textcolor{black}{standard deviation $\sigma=0.02$} (a3). Middle rows (b1-b3): true flow field (b1) and the mean flow field from the posterior distribution estimated using the MCMC samples. In the panels below (b2) and (b3), we also show the uncertainty regions (as shaded ellipses), also from MCMC samples as given by Eq.~(\ref{eq:cr}). Bottom row (c1-c3): time evolution (c1) as well as the distribution (c2-c3) of the effective regularization parameter $\sigma/\lambda$, where the distributions are obtained after discarding the initial transient in the MCMC sampling process.}
\label{fig2}
\end{figure}

\begin{figure}[htbp]
\centering
\includegraphics[width=0.75\textwidth]{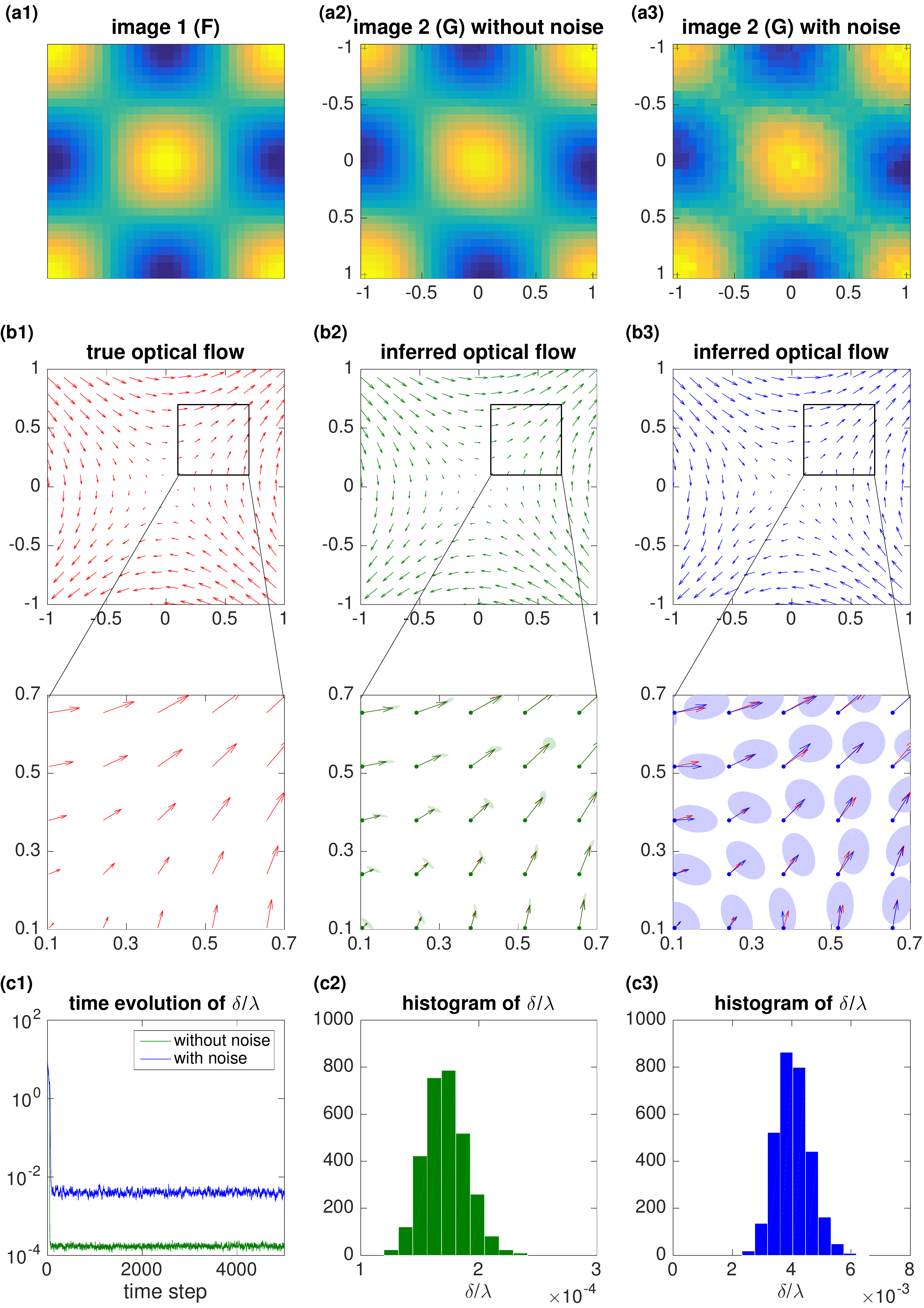}
\caption{Bayesian optical flow computed using statistical inversion, based on image pairs $(F,G)$ under example flow field 3, where $\langle U(x,y),V(x,y)\rangle = \langle y,\sin(x)\rangle$.  Top row (a1-a3): image data generated according to Eq.~(\ref{eq:hsexact}) for $F$ (a1) and using Eq.~(\ref{eq:firstimage}) for $G$ either without noise (a2) or with noise standard deviation $\sigma=0.02$ (a3). Middle rows (b1-b3): true flow field (b1) and the mean flow field from the posterior distribution estimated using the MCMC samples. In the panels below (b2) and (b3), we also show the uncertainty regions (as shaded ellipses), also from MCMC samples as given by Eq.~(\ref{eq:cr}). Bottom row (c1-c3): time evolution (c1) as well as the distribution (c2-c3) of the effective regularization parameter $\sigma/\lambda$, where the distributions are obtained after discarding the initial transient in the MCMC sampling process.}
\label{fig3}
\end{figure}

\begin{figure}[htbp]
\centering
\includegraphics[width=0.75\textwidth]{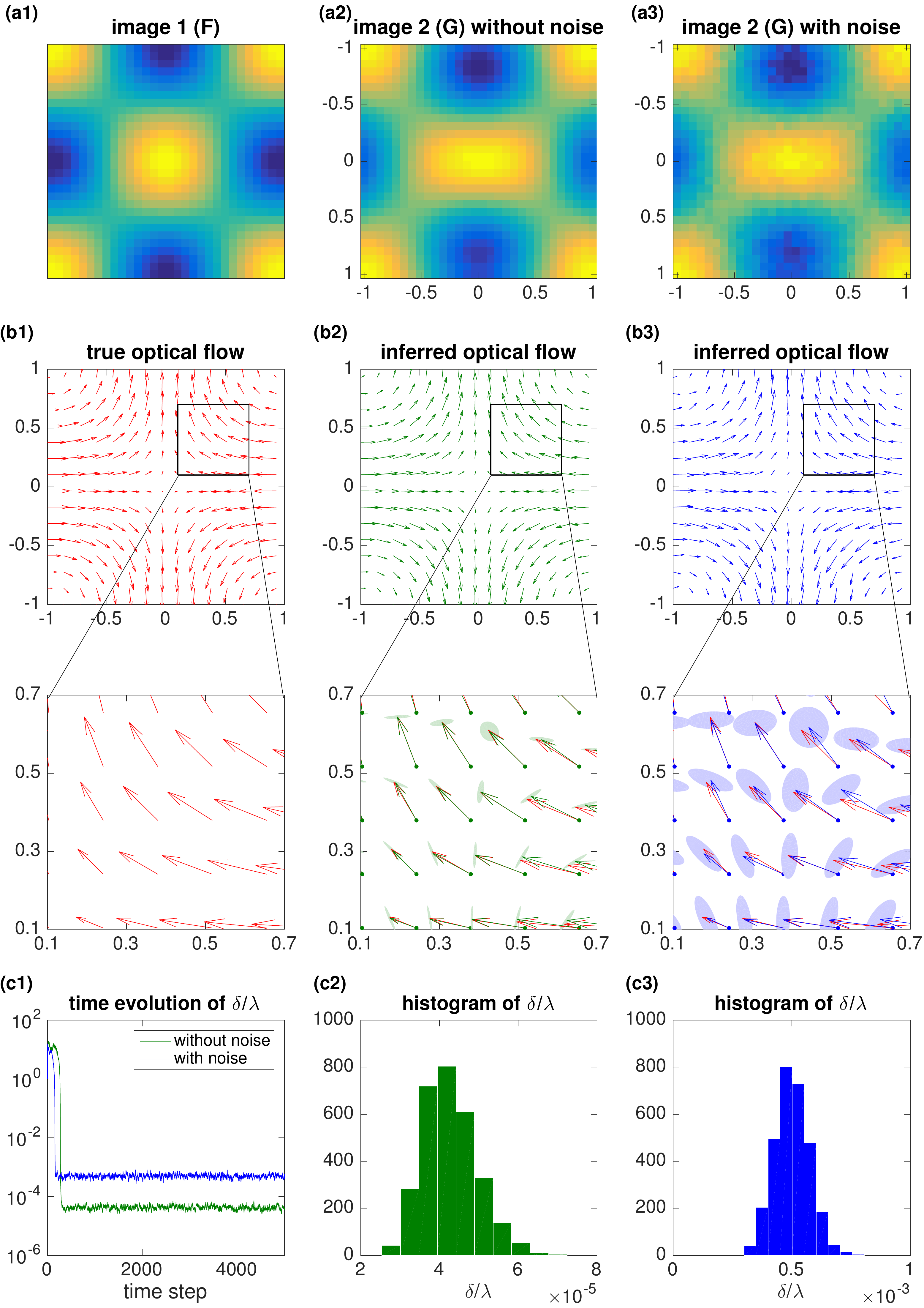}
\caption{Bayesian optical flow computed using statistical inversion, based on image pairs $(F,G)$ under example flow field 4, where $\langle U(x,y),V(x,y)\rangle = \langle -\pi\sin(0.5\pi x)\cos(0.5\pi y),\pi\cos(0.5\pi x)\sin(0.5\pi y)\rangle$.  Top row (a1-a3): image data generated according to Eq.~(\ref{eq:hsexact}) for $F$ (a1) and using Eq.~(\ref{eq:firstimage}) for $G$ either without noise (a2) or with noise standard deviation $\sigma=0.02$ (a3). Middle rows (b1-b3): true flow field (b1) and the mean flow field from the posterior distribution estimated using the MCMC samples. In the panels below (b2) and (b3), we also show the uncertainty regions (as shaded ellipses), also from MCMC samples as given by Eq.~(\ref{eq:cr}). Bottom row (c1-c3): time evolution (c1) as well as the distribution (c2-c3) of the effective regularization parameter $\sigma/\lambda$, where the distributions are obtained after discarding the initial transient in the MCMC sampling process.}
\label{fig4}
\end{figure}

\begin{figure}[htbp]
\centering
\includegraphics[width=0.75\textwidth]{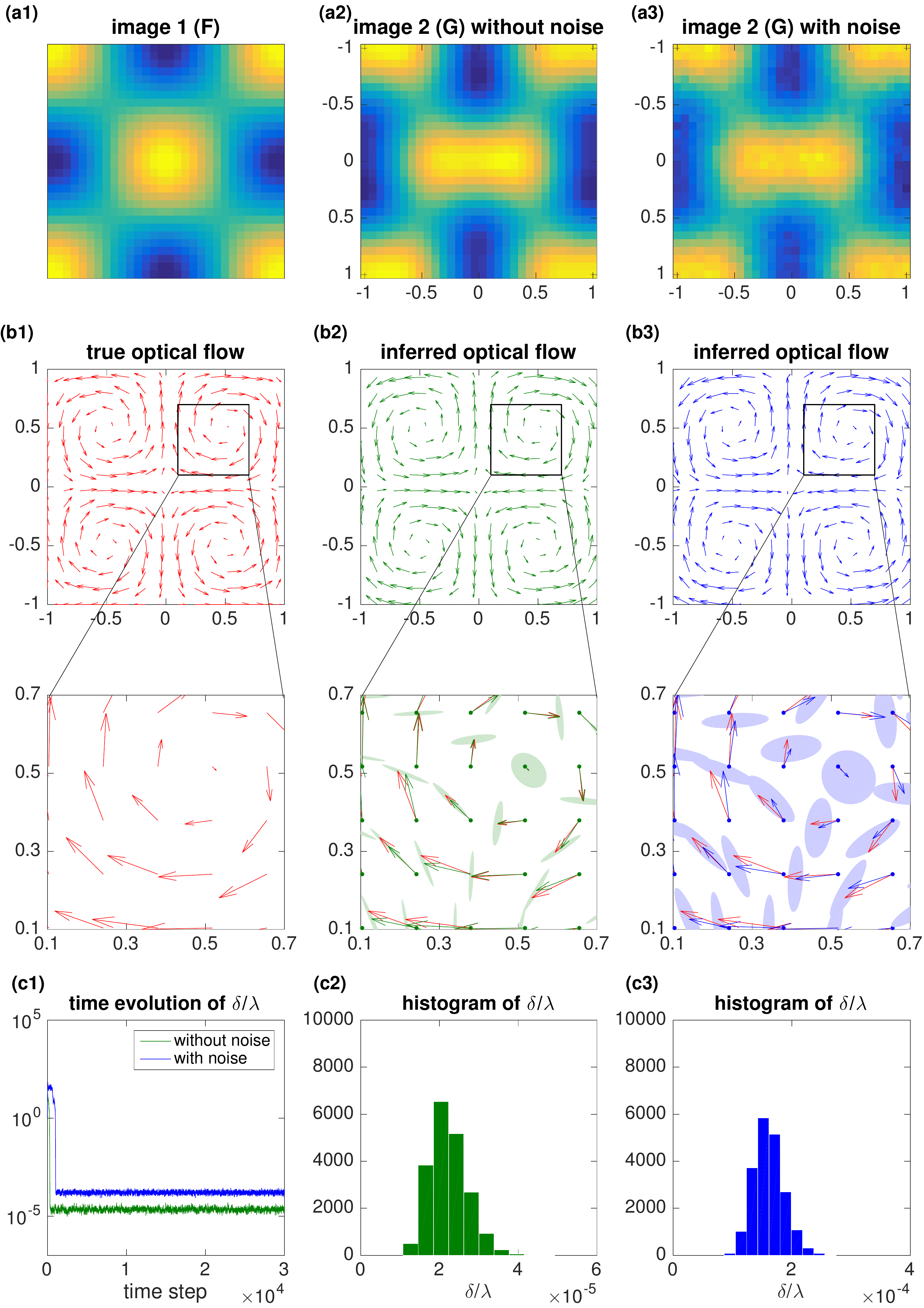}
\caption{Bayesian optical flow computed using statistical inversion, based on image pairs $(F,G)$ under example flow field 5, where $\langle U(x,y),V(x,y)\rangle = \langle -\pi\sin(\pi x)\cos(\pi y),\pi\cos(\pi x)\sin(\pi y)\rangle$. Top row (a1-a3): image data generated according to Eq.~(\ref{eq:hsexact}) for $F$ (a1) and using Eq.~(\ref{eq:firstimage}) for $G$ either without noise (a2) or with noise standard deviation $\sigma=0.02$ (a3). Middle rows (b1-b3): true flow field (b1) and the mean flow field from the posterior distribution estimated using the MCMC samples. In the panels below (b2) and (b3), we also show the uncertainty regions (as shaded ellipses), also from MCMC samples as given by Eq.~(\ref{eq:cr}). Bottom row (c1-c3): time evolution (c1) as well as the distribution (c2-c3) of the effective regularization parameter $\sigma/\lambda$, where the distributions are obtained after discarding the initial transient in the MCMC sampling process.}
\label{fig5}
\end{figure}

\begin{figure}[htbp]
\centering
\includegraphics[width=0.75\textwidth]{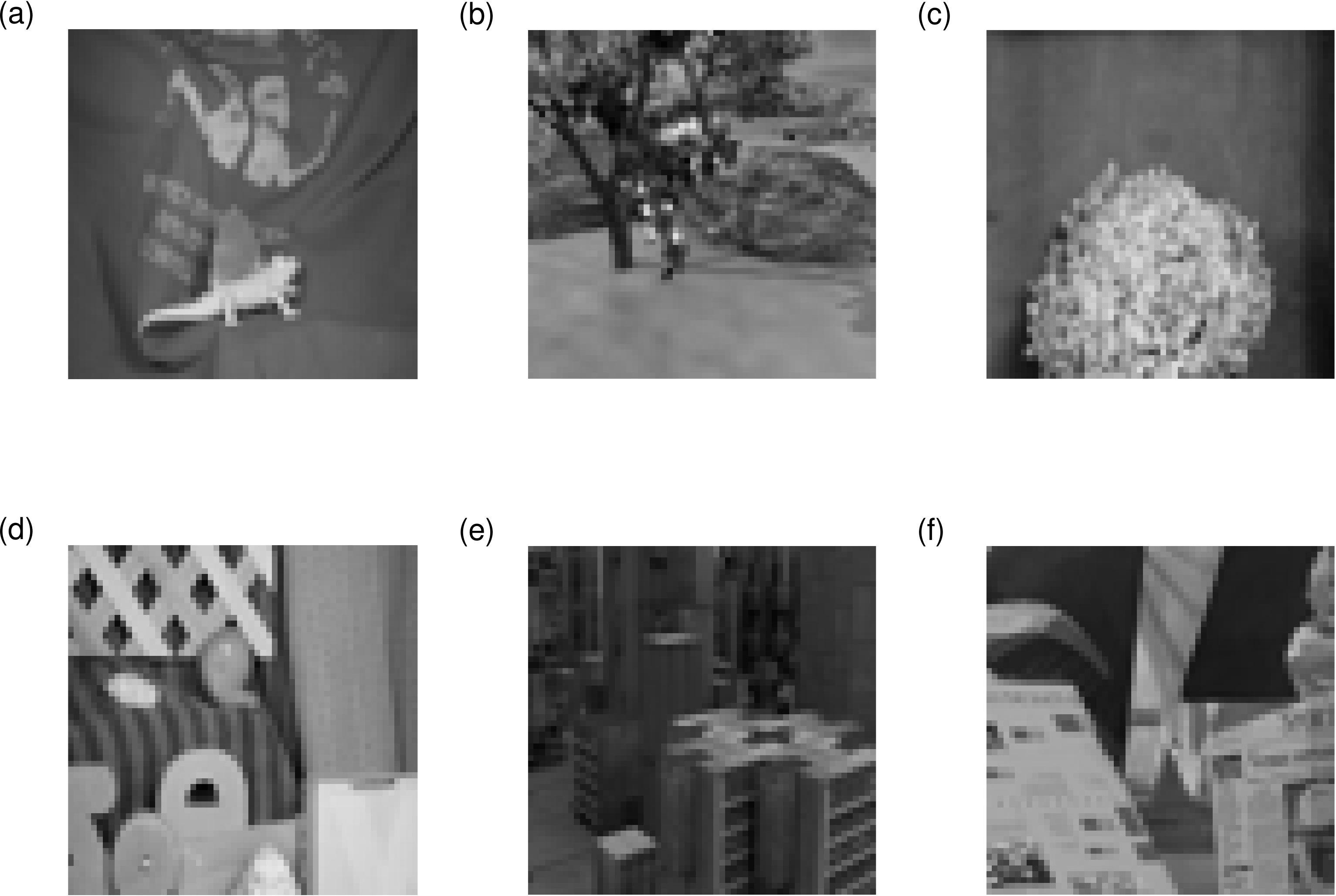}
\caption{Real images used for the computations of Bayesian optical flow in Sec.~\ref{sec:5.3}.}
\label{fig6}
\end{figure}

For each image pair $(F,G)$, we adopt the MCMC-Gibbs sampling procedure and corresponding choice of prior pdf and hyerpriors presented in Section 4 to obtain an empirical posterior distribution $p(U,V)$ as the solution of the statistical inversion problem, which we refer to as {\it Bayesian optical flow.} Unlike classical optical flow which provides a point estimate, the Bayesian optical flow can be thought of as an ensemble of flow fields each associated with some probability. From such an ensemble and associated probability distribution (the posterior distribution), we can further extract useful information. For example, the mean flow field can be computed using the sampling mean of the posterior distribution, which is shown in Fig.~\ref{fig1} through Fig.~\ref{fig6} to compare with the true underlying flow field.
In particular, in each figure, the top row (a1-a3) shows the image data of the first image $F$ (a1), and the second image $G$ generated from Eq.~(\ref{eq:hsexact}) with no noise (a2) and with noise under standard deviation $\sigma=0.02$ (a3), respectively.
The middle rows (b1-b3) show the true optical flow field (b1) compared with the inferred ``mean" optical flow fields together with uncertainty quantification from the MCMC samples (b2-b3). 
The uncertainty regions are computed as follows. At each point $z=(x,y)$, we construct a 2d normal pdf $\mathcal{N}(\mu,\Sigma)$ by using the sample mean $\mu$ and sample covariance $\Sigma$ estimated from the MCMC samples after discarding the initial transients. This allows us to obtain a ``mean" optical flow at point $z$ defined as $\langle u(z),v(z)\rangle=\langle \mu_1,\mu_2\rangle$. Uncertainty is quantified by computing a confidence region that contains $q$ probability mass of the fitted multivariate normal distribution given by~\cite{Siotani1964}
\begin{equation}\label{eq:cr}
	(z-\mu)^\top\Sigma^{-1}(z-\mu)\leq \chi_2^2(q).
\end{equation}
Here $\chi_2^2(q)$ denotes the $q$-th quantile of the Chi-squared distribution with two degrees of freedom, that is, $\chi_2^2(q)=K^{-1}(q)$ where $K$ is the cdf of $\chi_2^2$. These confidence regions (shaded ellipses) are shown for the zoomed-in plots for the inferred optical flow fields.
Finally, the last row (c1-c3) in each figure shows how the MCMC procedure produces a distribution of the effective regularization parameter $\delta/\lambda$. Panel (c1) shows the change of $\delta/\lambda$ over time in the MCMC sampling procedure, indicating convergence to a stationary distribution typically after a quick initial transient. The remaining panels (c2) and (c3) show the distribution of $\delta/\lambda$ after discarding the initial transient, for both the case of no noise (c2) and the case with noise (c3).

We point out a few observations from the numerical experiments. 
First, the estimated mean optical flow compare reasonably well with the true flow in all examples of the qualitatively different optical flow fields, supporting the utility of the proposed statistical inversion approach [see panels (b1-b3) in all figures].
Secondly, we again point out that the MCMC procedure used in our statistical inversion approach to optical flow does not require an active prior choice of the regularization parameter. The MCMC samples seems to quickly converge to a stationary distribution for the effective parameter [panel (c1) in all figures], from which the distributions of parameters and solutions can be determined.
Finally, comparing to the noise free images, the estimation of optical flow becomes less accurate when noise is added. It is worth mentioning that the statistical inversion approach in fact allows us to ``predict" this difference without knowing the ground-truth optical flow, by quantifying and comparing the uncertainty of solutions [panels (b2) and (b3) in all figures].
Note that although here we only show the Bayesian optical flow results based on Gaussian noise, we have also performed simulations using uniform noise and Laplace-distributed noise, and found that the results are quite similar to what is shown in Fig.~\ref{fig1} to Fig.~\ref{fig6}.

\subsection{Simulations on Real Images}\label{sec:5.3}
Finally, we conduct numerical experiments for the computation of Bayesian optical flow from using real images together with the benchmark flow fields. In each example, we consider matrix $F$ defined by a real image from the Middlebury dataset (\url{http://vision.middlebury.edu/stereo/}, also see~\cite{Sun2010}), resized to a fixed size of $60$-by-$60$ pixels in grayscale, with intensity normalized so that each pixel intensity is in the range $[0,1]$. We consider a total of six such images, as shown in Fig.~\ref{fig6}.

\begin{figure}[htbp]
\centering
\includegraphics[width=0.75\textwidth]{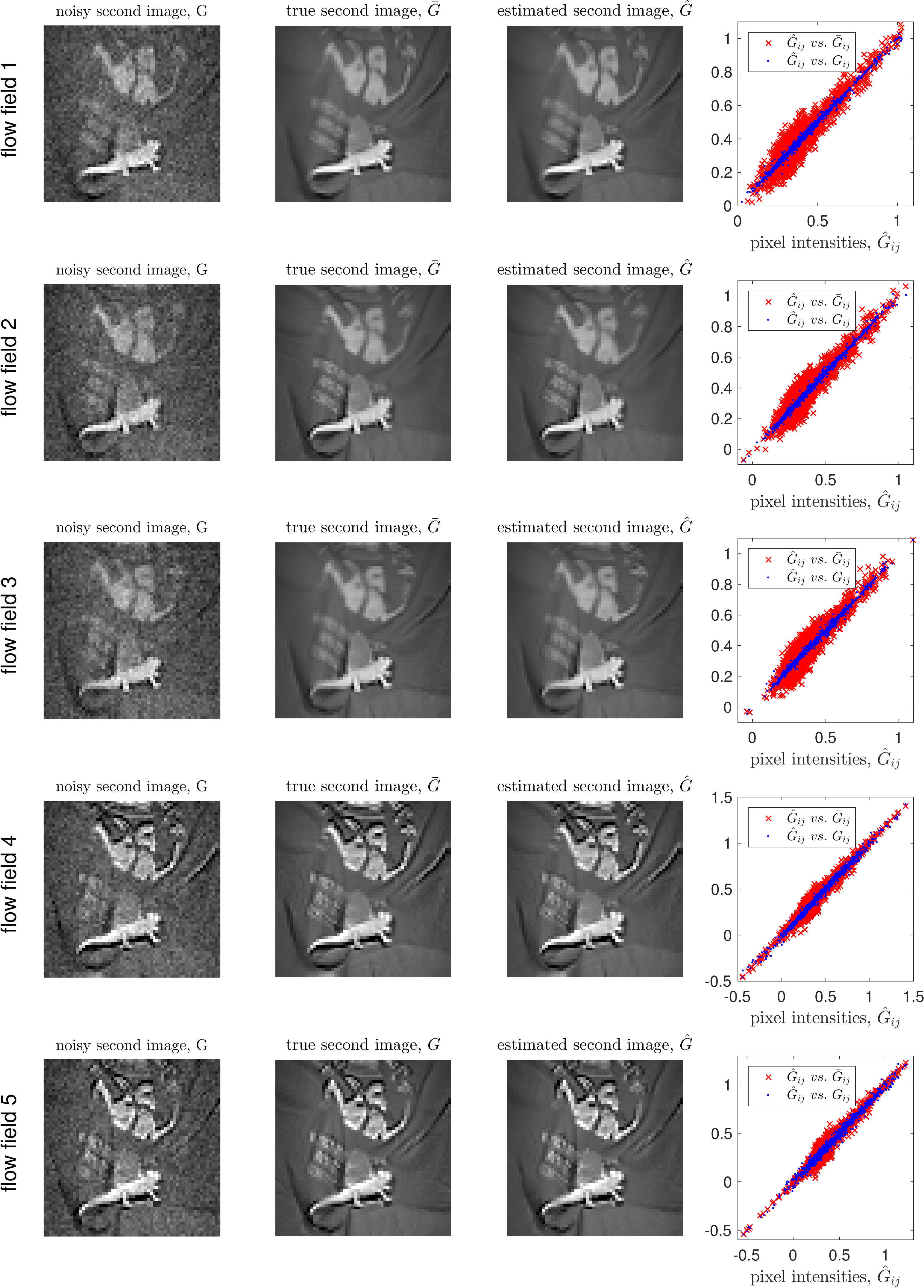}
\caption{Optical flow computed from noisy image pairs $(F,G)$, where $F$ is given by the image shown in Fig.~\ref{fig6}(a). Each row corresponds to the choice of a different flow field, each of which is defined in Sec.~\ref{sec:5.1}. From left to right in each row: noisy second image $G$ obtained from the flow equation~(\ref{eq:hsexact}) using Gaussian noise with $\sigma=0.05$, true second image $\bar{G}$ from the same flow equation in the absence of noise, estimated $\hat{G}$ based on using the mean flow field, and plots of $\hat{G}$ against $G$ versus $\hat{G}$ against $\bar{G}$.}
\label{fig7}
\end{figure}

\begin{figure}[htbp]
\centering
\includegraphics[width=0.75\textwidth]{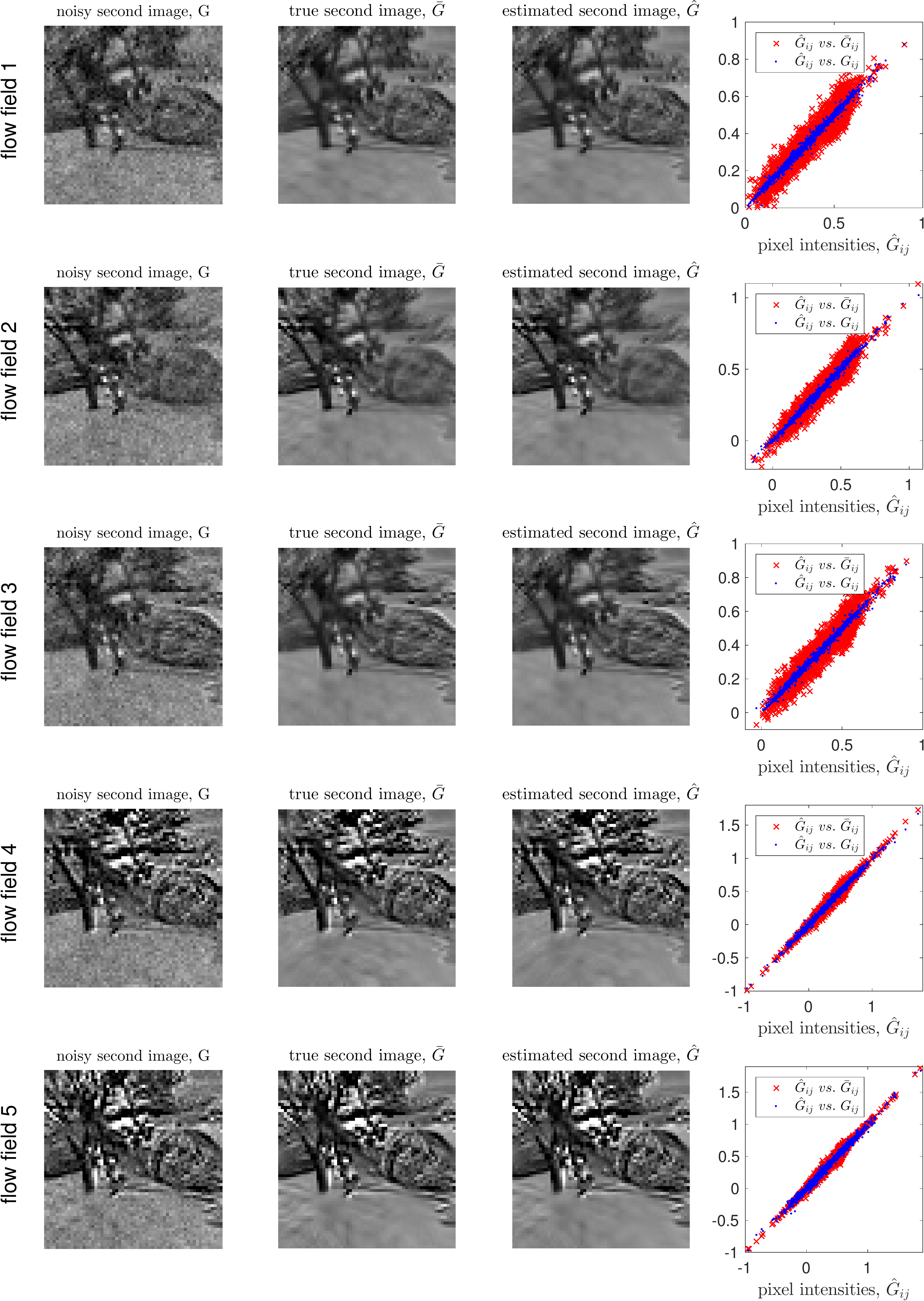}
\caption{Optical flow computed from noisy image pairs $(F,G)$, where $F$ is given by the image shown in Fig.~\ref{fig6}(b). Each row corresponds to the choice of a different flow field, each of which is defined in Sec.~\ref{sec:5.1}. From left to right in each row: noisy second image $G$ obtained from the flow equation~(\ref{eq:hsexact}) using Gaussian noise with $\sigma=0.05$, true second image $\bar{G}$ from the same flow equation in the absence of noise, estimated $\hat{G}$ based on using the mean flow field, and plots of $\hat{G}$ against $G$ versus $\hat{G}$ against $\bar{G}$.}
\label{fig8}
\end{figure}

\begin{figure}[htbp]
\centering
\includegraphics[width=0.75\textwidth]{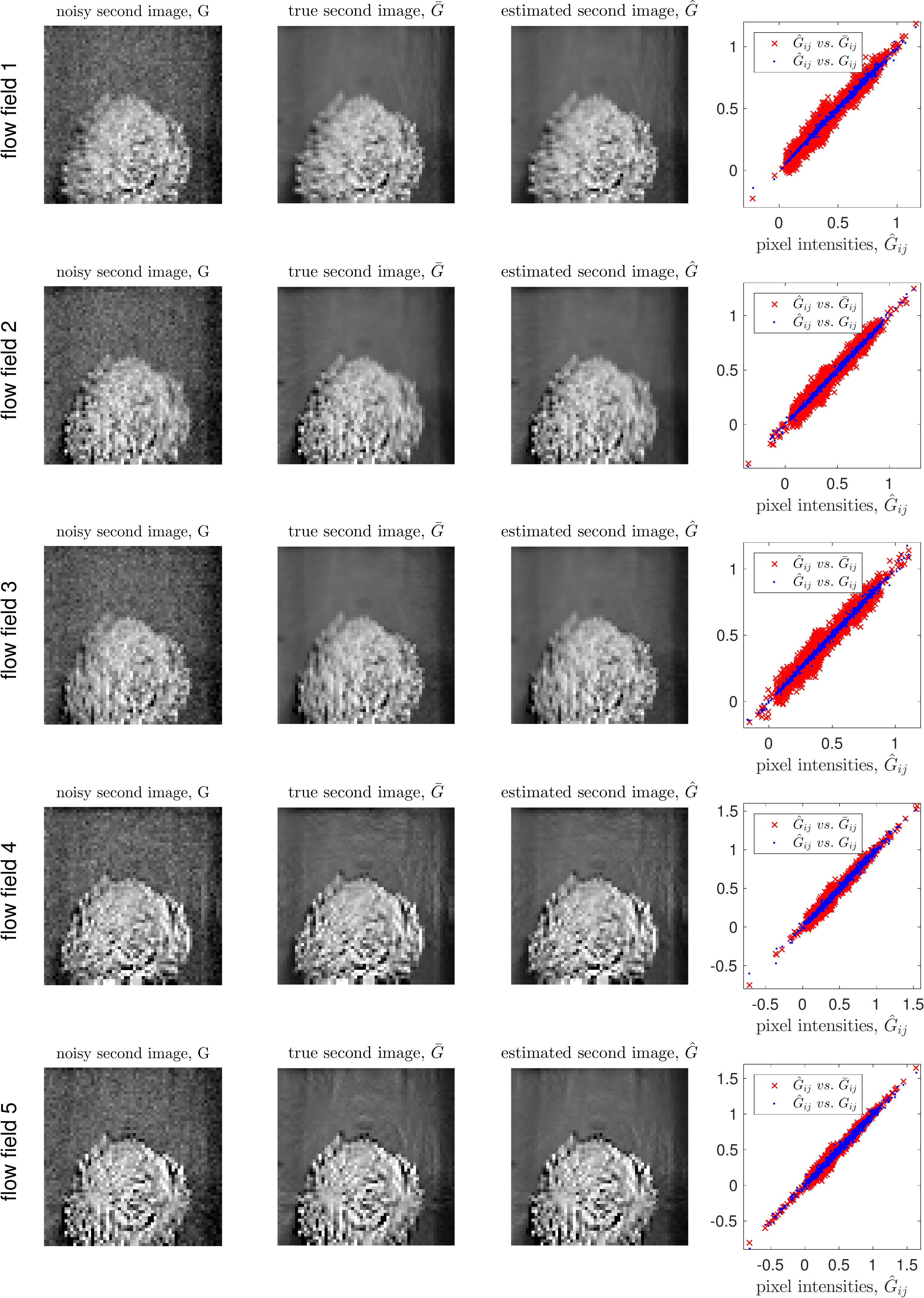}
\caption{Optical flow computed from noisy image pairs $(F,G)$, where $F$ is given by the image shown in Fig.~\ref{fig6}(c). Each row corresponds to the choice of a different flow field, each of which is defined in Sec.~\ref{sec:5.1}. From left to right in each row: noisy second image $G$ obtained from the flow equation~(\ref{eq:hsexact}) using Gaussian noise with $\sigma=0.05$, true second image $\bar{G}$ from the same flow equation in the absence of noise, estimated $\hat{G}$ based on using the mean flow field, and plots of $\hat{G}$ against $G$ versus $\hat{G}$ against $\bar{G}$.}
\label{fig9}
\end{figure}

\begin{figure}[htbp]
\centering
\includegraphics[width=0.75\textwidth]{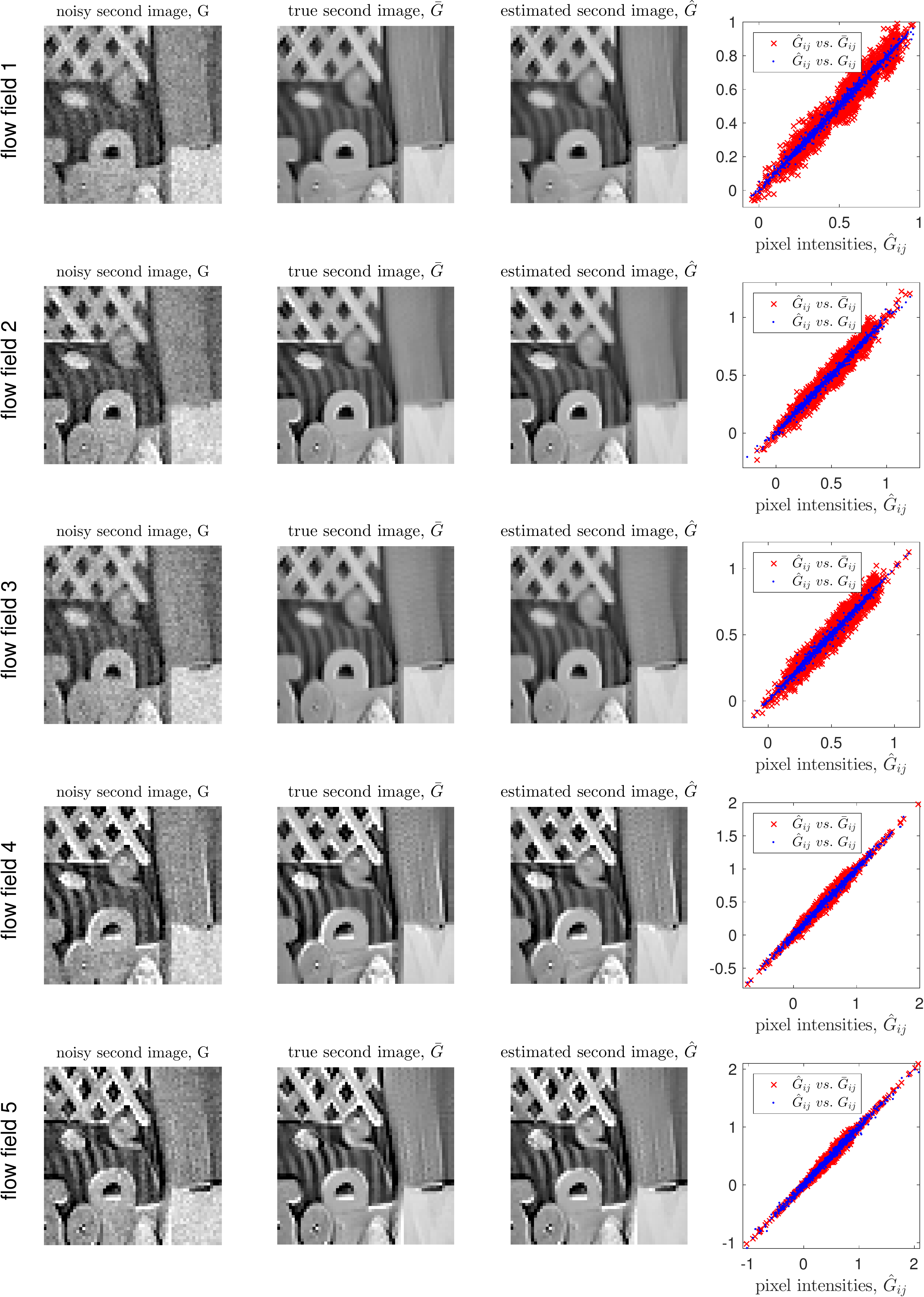}
\caption{Optical flow computed from noisy image pairs $(F,G)$, where $F$ is given by the image shown in Fig.~\ref{fig6}(d). Each row corresponds to the choice of a different flow field, each of which is defined in Sec.~\ref{sec:5.1}. From left to right in each row: noisy second image $G$ obtained from the flow equation~(\ref{eq:hsexact}) using Gaussian noise with $\sigma=0.05$, true second image $\bar{G}$ from the same flow equation in the absence of noise, estimated $\hat{G}$ based on using the mean flow field, and plots of $\hat{G}$ against $G$ versus $\hat{G}$ against $\bar{G}$.}
\label{fig10}
\end{figure}

\begin{figure}[htbp]
\centering
\includegraphics[width=0.75\textwidth]{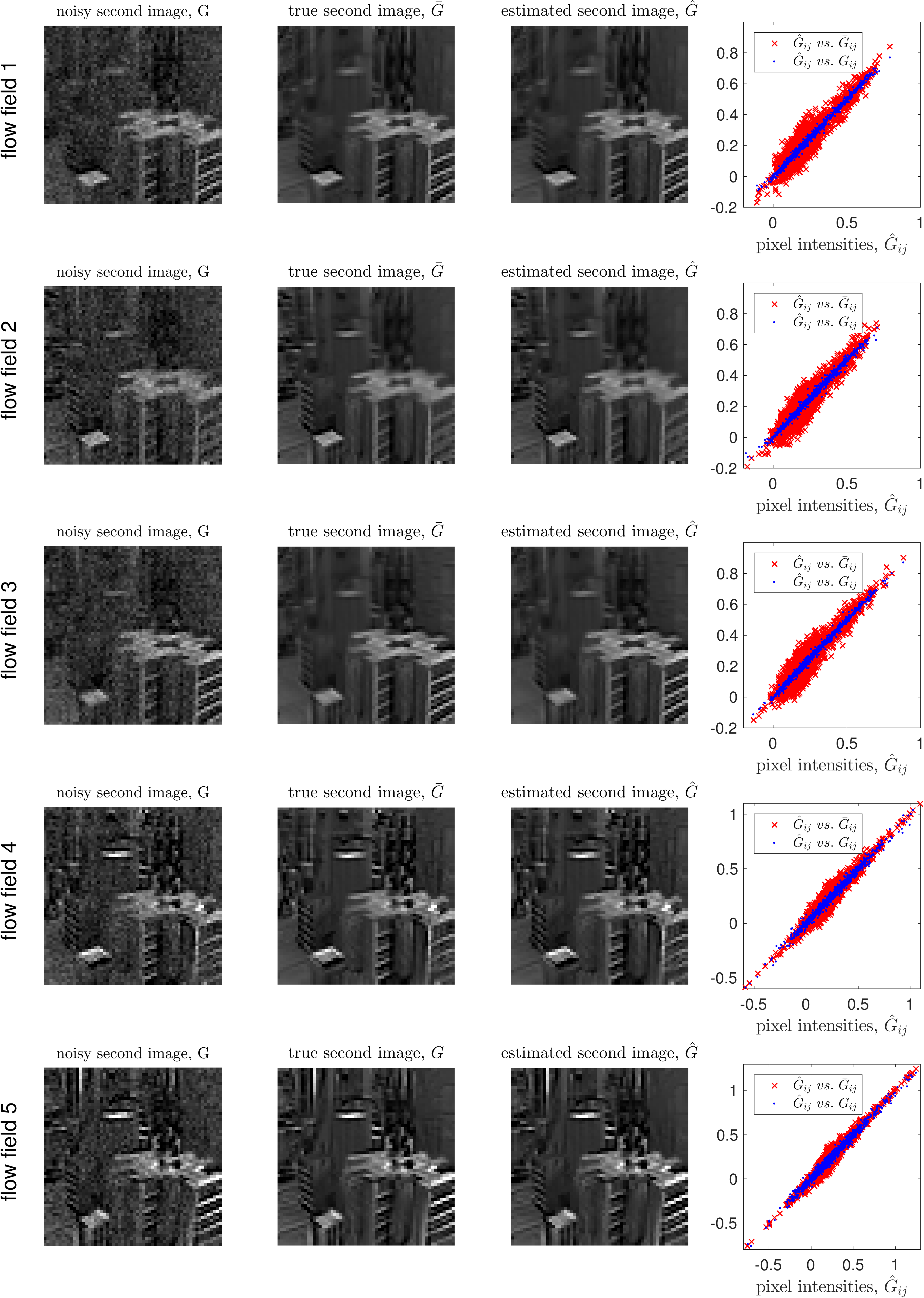}
\caption{Optical flow computed from noisy image pairs $(F,G)$, where $F$ is given by the image shown in Fig.~\ref{fig6}(e). Each row corresponds to the choice of a different flow field, each of which is defined in Sec.~\ref{sec:5.1}. From left to right in each row: noisy second image $G$ obtained from the flow equation~(\ref{eq:hsexact}) using Gaussian noise with $\sigma=0.05$, true second image $\bar{G}$ from the same flow equation in the absence of noise, estimated $\hat{G}$ based on using the mean flow field, and plots of $\hat{G}$ against $G$ versus $\hat{G}$ against $\bar{G}$.}
\label{fig11}
\end{figure}

\begin{figure}[htbp]
\centering
\includegraphics[width=0.75\textwidth]{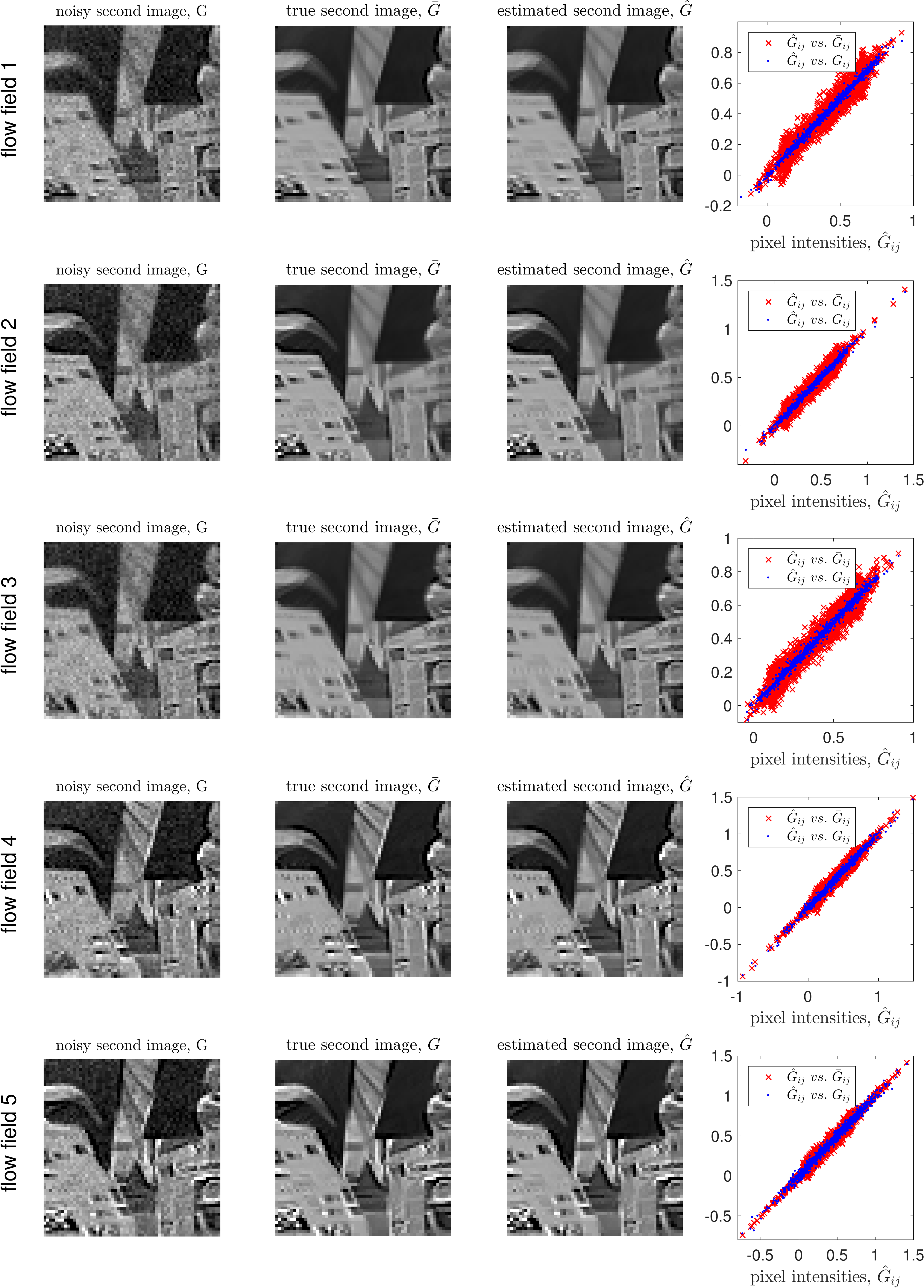}
\caption{Optical flow computed from noisy image pairs $(F,G)$, where $F$ is given by the image shown in Fig.~\ref{fig6}(f). Each row corresponds to the choice of a different flow field, each of which is defined in Sec.~\ref{sec:5.1}. From left to right in each row: noisy second image $G$ obtained from the flow equation~(\ref{eq:hsexact}) using Gaussian noise with $\sigma=0.05$, true second image $\bar{G}$ from the same flow equation in the absence of noise, estimated $\hat{G}$ based on using the mean flow field, and plots of $\hat{G}$ against $G$ versus $\hat{G}$ against $\bar{G}$.}
\label{fig12}
\end{figure}

For each real image $F$ and a given flow field $\langle U,V\rangle$, we generate a resulting second image $G$ according to the flow equation~(\ref{eq:hsexact}), where the noise is taken to be a multivariate Gaussian distribution with zero mean and covariance matrix $\sigma I$ with $\sigma=0.05$. This choice of standard deviation ensures a non-negligible effect on the pixel intensities of the image pairs, since each $F_{ij}\in[0,1]$. These ``noisy second images" are shown in each of the first column of Fig.~\ref{fig7} to Fig.~\ref{fig12}. For comparison, we also consider what we call a ``true second image", denoted as $\bar{G}$, which is defined by the same flow equation~(\ref{eq:hsexact}) but in the absence of noise. These true second images are shown in the second column in Fig.~\ref{fig7} through Fig.~\ref{fig12}, for each one of the image example and flow field. Thus, each baseline image $F$ and flow field gives rise to a noisy image pair $(F,G)$, and there is a total of 30 such pairs given the 6 real images and 5 flow fields.

For each noisy image pair $(F,G)$, we use the same methodology with the same choice of priors and hyperpriors as in Sec.~\ref{sec:5.2} to obtain a Bayesian optical flow from sampling the posterior distribution--in practice because of the randomness of the MCMC process, we observe that sometimes the sample distribution does not ``converge" to a stationary distribution even after thousands of iterations; when this occurs we simply restart the MCMC from a different initial seed. Recall that different from a standard regularization approach which require careful choice of the regularization parameter, here such parameter is itself treated as a random variable that has its own prior which can be taken to be a ``wide" distribution in the absence of additional knowledge. From the posterior distribution, we take the mean as an estimate of the average flow field. To test the usefulness of the reconstructed flow field, we use it together with the first image $F$ to obtain an estimated second image $\hat{G}$ from equation~(\ref{eq:hsexact}), setting the noise term to be zero. The estimated second image $\hat{G}$ is shown for each example in the third column of Fig.~\ref{fig7} through Fig.~\ref{fig12}. Interestingly, the estimated $\hat{G}$ seems to not only resemble the given noisy image $G$, but even more similar to the noiseless second image $\bar{G}$. This observation is confirmed quantitatively, as shown in the last column of Fig.~\ref{fig7} through Fig.~\ref{fig12}, where the values of $\hat{G}_{ij}$ are plotted against both those of $G_{ij}$ and of $\bar{G}_{ij}$. These results confirm the validity of Bayesian optical flow obtained by a statistical inversion approach, and, additional suggests that accurate reconstruction of optical flow can be potentially useful for image smoothing and denoising applications.

\section{Discussion and Conclusions}
In this paper we take a statistical inversion perspective to the optical flow inference problem. From this perspective all relevant variables in an otherwise standard inverse problem are treated as random variables, and the key is to form an efficient process to construct and sample the posterior distribution by utilizing knowledge about the form of model, noise, and other prior information.   
From a Bayesian perspective, the various priors and the forward model combine to produce a posterior distribution describing the propagation of prior information in context of the problem. We have shown that optical flow estimates given by traditional variational approaches such as the seminal work developed by Horn and Schunck~\cite{HS1981} can in fact be interpreted in the statistical inversion framework under particular choices of models, noise, and priors. Thus we recap that there are major advantages over the classical variational calculus approach to inverse problems where by necessity the ill-posedness is dealt with by adding an ad hoc regularity term that hopefully agrees with expected physical interpretation.  From a Bayesian perspective, the ill-posedness is dealt with naturally under the statistical inversion framework by restating as a well-posed extended problem in a larger space of probability distributions~\cite{GelmanBook,KaipioBook}.  This therefore naturally removes a key difficulty of having to choose exactly an appropriate regularity parameter encountered in classical methods. Instead, in contrast to classical optical flow methods which only yield single solutions as ``point estimates", the statistical inversion approach produces a distribution of points which can be sampled in terms of most appropriate estimators and also for uncertainty quantification.  Specifically in the context of an optic flow problem, we expect a distribution of regularity parameters, and correspondingly a distribution of optical flow vectors at each pixel.

In this paper we focused on a statistical inversion formulation with the choice of priors inspired from the classical Horn-Schunck functional. Although such choices appear to be reasonable for the synthetic flow fields considered herein, real images and optical flow fields are much more complicated and, consequently, require additional efforts in forming the appropriate priors. Such priors can come from previous knowledge of images and flow fields from similar systems, taken under similar scenes, or from other physical measurements. In addition, different noise distributions should be considered, and these will require modifying the standard Horn-Schunck framework, which assumed rigid body motion and conservation of brightness. In particular, with regard to the regularization term, one example is to consider different $p$-norms other than the standard $2$-norm, and possibly with a kernel to weigh in additional information about the physical embedding of objects~\cite{Sun2017}.
Other data fidelity terms~\cite{Basnayake2014, lutt, bays} can be formulated to describe the physics of the underlying application, for example for fluid and oceanographic problems where a stream function, or even a quasi-static approximation to assume such physics as corriolis can be used; or in scenarios where divergence-free flows are estimated by utilizing vorticity-velocity formalism or more general data assimilation tools~\cite{Bereziat2011,Herlin2012,Zhuk2017}. Likewise, regularity in time and multiple time steps may be appropriate~\cite{Basnayake2014}, as these correspondingly more complex formulations nonetheless come back to a linear inverse problem tenable in the framework of this paper.   Specifically within the statistical inversion framework developed in the current paper, all of these could be recast so that the data fidelity term may be incorporated and should not require significant modification of the general framework as introduced here, which we plan in future work.  Likewise, other numerical differentiation and integration schemes can be used as well in place of the simple forward difference used here.


\newpage

\newpage

\end{document}